\documentclass[runningheads]{llncs}
\usepackage{graphicx}
\usepackage{comment}
\usepackage{amsmath,amssymb} 
\usepackage{color}

\usepackage{subcaption}
\usepackage{lipsum}
\usepackage[table]{xcolor}
\usepackage{booktabs}

\usepackage{multirow}

\usepackage{cite}

\begin{document}
\pagestyle{headings}
\mainmatter
\def\ECCVSubNumber{2847}  

\title{Extended Labeled Faces in-the-Wild (ELFW): \\Augmenting Classes for Face Segmentation}


\titlerunning{Extended Labeled Faces in the Wild (ELFW)}
\author{Rafael Redondo\inst{1}\and Jaume Gibert\inst{1}}
\authorrunning{R. Redondo and J. Gibert}

\institute{Eurecat, Centre Tecnològic de Catalunya, Multimedia Technologies, Barcelona, Spain
\email{\{rafael.redondo,jaume.gibert\}@eurecat.org}\\
\url{https://multimedia-eurecat.github.io}}


\newcommand{\figrowsep}{\vspace{-1.0pt}}
\newcommand{\figcolsep}{\hspace{0.0pt}}
\newcommand{\captionsep}{\vspace{-12pt}}
\newcommand{\ie}{\textit{i.e.}\ }
\newcommand{\eg}{\textit{e.g.}\ }


\maketitle

\begin{abstract}

Existing face datasets often lack sufficient representation of occluding objects, which can hinder recognition, but also supply meaningful information to understand the visual context.
In this work, we introduce Extended Labeled Faces in-the-Wild (ELFW)\footnote{ELFW dataset and code can be downloaded from~\url{https://multimedia-eurecat.github.io/2020/06/22/extended-faces-in-the-wild.html}.}, a dataset supplementing with additional face-related categories ---and also additional faces--- the originally released semantic labels in the vastly used Labeled Faces in-the-Wild (LFW) dataset.
Additionally, two object-based data augmentation techniques are deployed to synthetically enrich under-represented categories which, in benchmarking experiments, reveal that not only segmenting the augmented categories improves, but also the remaining ones benefit.

\end{abstract}

\begin{figure}[t]
\centering
\begin{subfigure}[b]{0.5879\textwidth}
\includegraphics[width=0.19\linewidth]{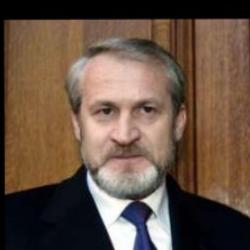}\figcolsep
\includegraphics[width=0.19\linewidth]{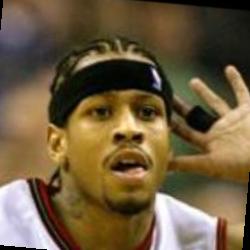}\figcolsep
\includegraphics[width=0.19\linewidth]{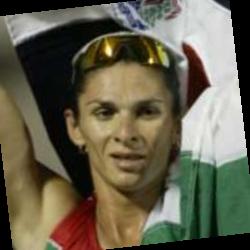}\figcolsep
\includegraphics[width=0.19\linewidth]{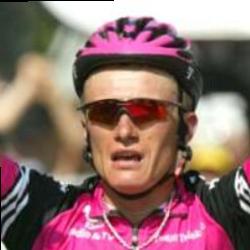}\figcolsep
\includegraphics[width=0.19\linewidth]{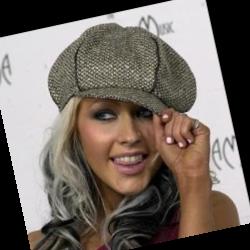}\figcolsep

\figrowsep

\includegraphics[width=0.19\linewidth]{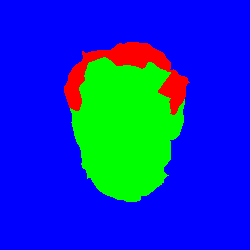}\figcolsep
\includegraphics[width=0.19\linewidth]{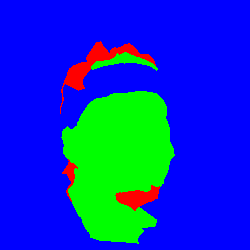}\figcolsep
\includegraphics[width=0.19\linewidth]{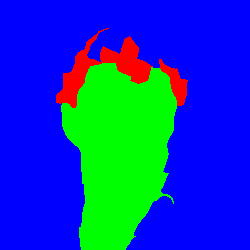}\figcolsep
\includegraphics[width=0.19\linewidth]{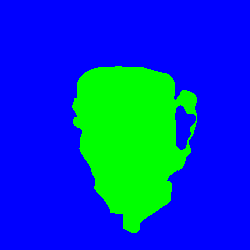}\figcolsep
\includegraphics[width=0.19\linewidth]{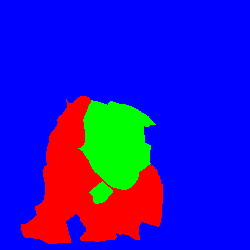}\figcolsep

\figrowsep

\includegraphics[width=0.19\linewidth]{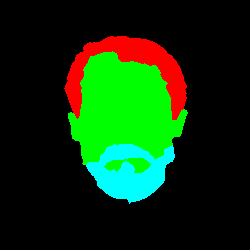}\figcolsep
\includegraphics[width=0.19\linewidth]{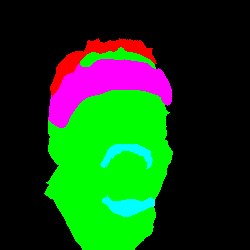}\figcolsep
\includegraphics[width=0.19\linewidth]{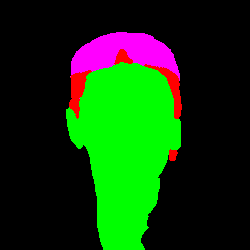}\figcolsep
\includegraphics[width=0.19\linewidth]{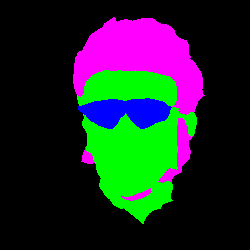}\figcolsep
\includegraphics[width=0.19\linewidth]{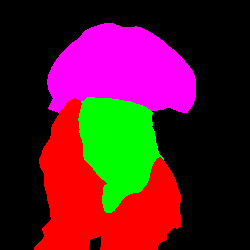}\figcolsep
\end{subfigure}
\hspace{-1.6em}
\begin{subfigure}[b]{0.438\textwidth}
\includegraphics[width=0.19\linewidth]{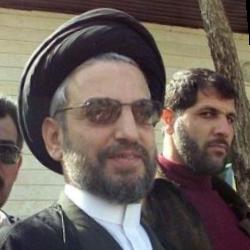}\figcolsep
\includegraphics[width=0.19\linewidth]{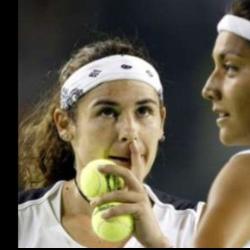}\figcolsep
\includegraphics[width=0.19\linewidth]{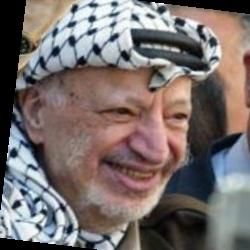}\figcolsep
\includegraphics[width=0.19\linewidth]{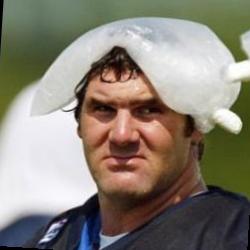}\figcolsep
\includegraphics[width=0.19\linewidth]{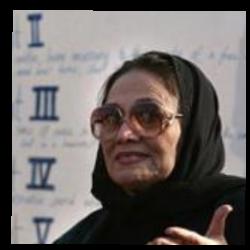}\figcolsep

\figrowsep

\includegraphics[width=0.19\linewidth]{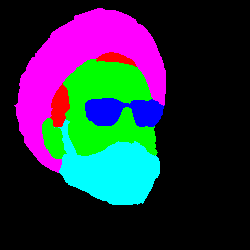}\figcolsep
\includegraphics[width=0.19\linewidth]{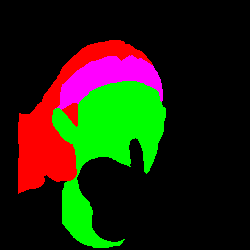}\figcolsep
\includegraphics[width=0.19\linewidth]{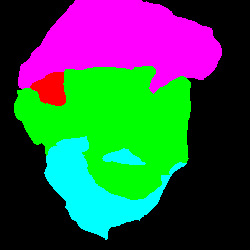}\figcolsep
\includegraphics[width=0.19\linewidth]{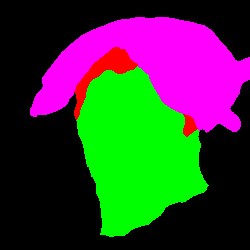}\figcolsep
\includegraphics[width=0.19\linewidth]{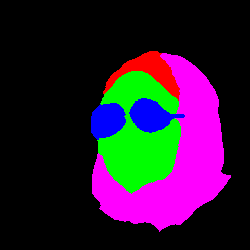}\figcolsep

\includegraphics[width=0.19\linewidth]{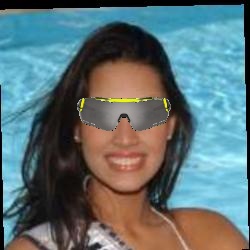}\figcolsep
\includegraphics[width=0.19\linewidth]{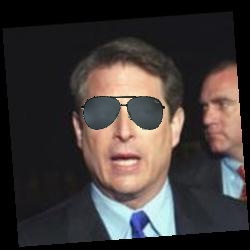}\figcolsep
\includegraphics[width=0.19\linewidth]{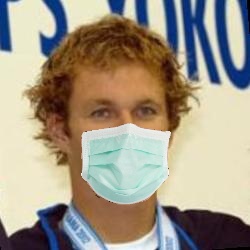}\figcolsep
\includegraphics[width=0.19\linewidth]{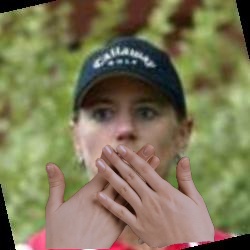}\figcolsep
\includegraphics[width=0.19\linewidth]{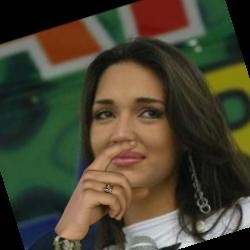}\figcolsep

\figrowsep

\includegraphics[width=0.19\linewidth]{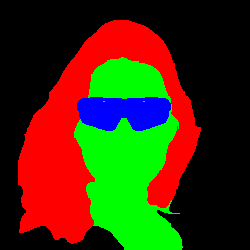}\figcolsep
\includegraphics[width=0.19\linewidth]{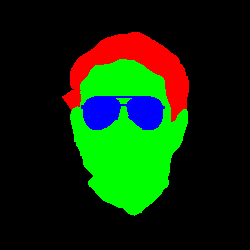}\figcolsep
\includegraphics[width=0.19\linewidth]{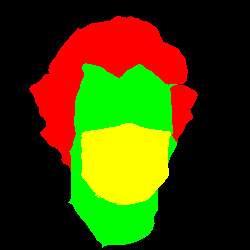}\figcolsep
\includegraphics[width=0.19\linewidth]{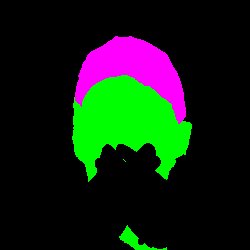}\figcolsep
\includegraphics[width=0.19\linewidth]{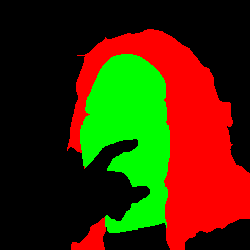}\figcolsep
\end{subfigure}

\caption{Examples of the ELFW dataset: the original LFW categories \textit{background}, \textit{skin}, and \textit{hair}, the new categories \textit{beard-mustache}, \textit{sunglasses}, \textit{head-wearable}, and the exclusively synthetic \textit{mouth-mask}. (left) Re-labeled faces with manual refinement compared to the original LFW labels in blue background, (right-top) faces previously unlabeled in LFW, and (right-bottom) synthetic object augmentation with sunglasses, mouth-masks, and occluding hands.}
\label{fig:elfw_examples}
\end{figure}

\section{Introduction}



The existence of convenient datasets is of paramount importance in the present deep learning era and the domain of facial analysis is not exempt from this situation. Data should not only be accurate, but also large enough to describe all those underlying features fundamental for machine learning tasks. The first face datasets were acquired under controlled conditions~\cite{phillips2005overview, jesorsky2001robust, phillips1998feret, sim2001cmu}, specially in relation to lightning, background, and facial expression. However, real-world applications in-the-wild operate under immeasurable conditions, which are often unrepeatable or at least an arduous task to be reproduced in the laboratory. In this respect, natural datasets acquired in-the-wild enable to cover a larger diversity, also with less effort.

Labeled Faces in-the-Wild (LFW)~\cite{LFWTech, huang2008labeled}, previous to the deep learning uprise, and a cornerstone to the present work, aimed at providing a large-scale face dataset to leverage applications in unconstrained environments, which are characterized by having extreme variations in image quality such as contrast, sharpness, or lighting, but also content variation such as head pose, expression, hairstyle, clothing, gender, age, race, or backgrounds. Additionally within LFW, semantic segmentation maps (labels) were released for a subset of faces with three different categories: \textit{background}, \textit{skin}, and \textit{hair}.
However, these maps lack contextual information given by either complementing or occluding objects and, particularly, they present several segmentation inconsistencies and inaccuracies. More specifically, beards and moustaches were inconsistently annotated either as skin or hair; any type of object on the head which obstacles identification is considered as part of the background, causing unnatural discontinuities in the facial semantic interpretation; common objects like sunglasses are simply ignored; and last but not least, not a few number of cases have irregular labeled boundaries due to the used super-pixel based annotation strategy.

The main goal of this work is to create a set of renewed labeled faces by improving the semantic description of objects commonly fluttering around faces in pictures, and thus enabling a richer context understanding in facial analysis applications.
To this end, we aim at extending LFW with more semantic categories and also more labeled faces, which partially solve the original LFW flaws and expand its range to a larger and more specialized real-life applications.
In particular, as illustrated in Fig.~\ref{fig:elfw_examples}, we extended the LFW dataset in three different ways:
(1) we updated the originally labeled semantic maps in LFW with new categories and refined contours, (2) we manually annotated additional faces not originally labeled in LFW, and (3) we automatically superimposed synthetic objects to augment under-represented categories in order to improve their learning.
With such an extension, we additionally provide results for state of the art baseline segmentation models to be used for future reference and evaluate these data augmentation techniques.


\newpage

\section{Related Works}

The LFW dataset is contemporary to several other datasets with a similar goal, \ie providing vast and rich facial attributes in the form of images and labels. Examples are the Helen dataset~\cite{le2012interactive}, with $2,330$ dense landmark annotated images, the Caltech Occluded Faces in-the-Wild (COFW)~\cite{burgos2013robust}, comprising landmark annotations on $1,007$ images of occluded faces, and the Caltech 10000 Web Faces~\cite{fink2007caltech}, a larger dataset designed for face detection in-the-wild, which however lacks aligned faces.

With time, datasets have grown larger and richer in attributes. One of the first widely-used public datasets for training deep models was the CASIA-Webface \cite{yi2014learning}, with $500K$ images of $10K$ celebrities. 
Subsequently, the IARPA Janus Benchmark and successive upgrades (NIST IJB-A,B,C)~\cite{klare2015pushing,whitelam2017iarpa,maze2018iarpa} released large datasets constructed upon still images and video frames, which were especially designed to have a more uniform geographic distribution. In such works, the authors claimed that the main limitation of previous databases, such as the Youtube Faces~\cite{wolf2011face,ferrari2018extended}, the really mega in terms of quantity MegaFace~\cite{Nech2017LevelPF}, and the really wide in terms of variety and scale WiderFace~\cite{yang2016wider}, is that they were constructed with basic face detectors, and thus rejecting many valid faces due to far-reaching perspectives and facial expressions. It is also noted that benchmarks in datasets such as LFW are saturated, where the best face recognition performance exceeds a $99\%$ true positive rate, which suggests the need for an  expansion of datasets. Another recent and also very large dataset is VggFace2~\cite{cao2018vggface2}, which took advantage of internet image search tools to download a huge number of images to later manually screen false positives. Also recently, the large-scale celebrity faces attributes CelebA~\cite{liu2018large} gathered over $200K$ aligned images of more than $10K$ celebrities, including $5$ landmarks and $40$ tagged attributes such as wearing hat, moustache, smiling, wavy hair, and others. Finally, the prominent MS-Celeb-1M~\cite{guo2016ms} ---recently shut down in response to journal investigations--- got to collect over $10$ million images from more than $100K$ celebrities.
Regarding components of heritage and individual identity reflected in faces, it is worth noting an outstanding study on diversity~\cite{merler2019diversity}, which provided 1 million annotated faces by means of coding schemes of intrinsic facial descriptions, mainly intended for face recognition.

Until now, face datasets for semantic segmentation were mostly focused on facial parts such as lips, eyebrows, or nose~\cite{warrell2009labelfaces,liu2015multi}. At the time this work was carried out, a notorious CelebAMaskHQ~\cite{lee2019maskgan} was released with $21$ categories over $30K$ high-quality images. Nonetheless, the labeling and detection of common occluding objects was out of their scope, but they can be certainly deemed as complementary to the present work. 
Considering that face recognition or face synthesis are not the final goal here, LFW was a good candidate which still offers a great variability to build upon. Moreover, it already provides pre-computed annotation segments and has been also heavily reported. The above mentioned limitations, though, should be considered when deploying applications for real environments.

\subsection{Labeled Faces in-the-Wild}
\label{sec:lfw_relatedwork}

Labeled Faces in-the-Wild (LFW) was originally created in the context of human face recognition, this is, the identification of particular individual faces. 
Although it is arguably an old dataset given the effervescent deep learning expansion, it has been regularly applied in numerous machine learning applications on computer vision. Examples are face verification~\cite{taigman2014deepface, schroff2015facenet, parkhi2015deep, wen2016discriminative, cao2010face, sun2015deepid3, hu2014discriminative, kumar2011describable, amos2016openface, liu2017sphereface}, high level features for image recognition~\cite{le2013building, liu2015deep, kumar2009attribute, sun2014deep, sun2015deeply, li2013learning, kumar2011describable, huang2012learning, bourdev2011describing, wolf2010effective, berg2013poof}, large scale metrics learning~\cite{koestinger2012large, coates2013deep, simonyan2013fisher, nguyen2010cosine}, face alignment~\cite{cao2014face, peng2012rasl, saragih2009face}, landmark and facial parts detection~\cite{sun2013deep, belhumeur2013localizing, ranjan2017hyperface}, generic image classification and similarity metrics~\cite{chan2015pcanet, larsen2015autoencoding, mignon2012pcca}, image retrieval~\cite{wan2014deep, kumar2011describable, siddiquie2011image}, age and gender classification~\cite{levi2015age, ranjan2017hyperface, eidinger2014age}, pose, gesture and gaze recognition~\cite{dantone2012real, ranjan2017hyperface, rivera2012local, zhu2015high, zhang2015appearance}, face frontalization~\cite{hassner2015effective}, and model warping~\cite{saragih2011deformable}, to name `a few'. 
Furthermore, other datasets have also been derived from it~\cite{koestinger2011annotated}.

The LFW dataset is made up of $13,233$ jpeg images of $250\times 250$ pixels from $5,749$ people, where $1,680$ people have two or more images\footnote{Varied erratas have been later published in consecutive amendments. For those meticulous readers, please visit the official website at~\url{http://vis-www.cs.umass.edu/lfw/index.html}.}. 
The authors advice that LFW has its own bias, as any other dataset. In particular, few faces present poor lightning exposure conditions and most pictures contain frontal portraits, because the Viola-Jones face detector~\cite{viola2004robust}, used for filtering --and cropping and resizing-- fails on angular views, highly occluded faces, and distant individuals.

LFW comes with different parallel datasets based on different alignment approaches: (1) the funneling method proposed by Huand et al.~\cite{huang2007unsupervised}, (2) the \textit{LFW-a}\footnote{\url{https://talhassner.github.io/home/projects/lfwa/index.html}} commercial software, and (3) deep funneled~\cite{Huang2012a}. Among these, the last two are claimed to provide superior results for most face verification algorithms. Additionally, all parallel versions have computed superpixel representations with the Mori's online implementation\footnote{\url{http://www.cs.sfu.ca/~mori/research/superpixels}}, an automatic local segmentation based on local color similarity~\cite{mori2005guiding}. Finally, by means of these superpixels, LFW released $2,927$ face images originally labeled with 3 categories: \textit{hair}, \textit{skin}, and \textit{background}.


\section{Extended Labeled Faces in-the-Wild}

\begin{figure}[t]
\begin{center}
	\includegraphics[width=0.7\linewidth]{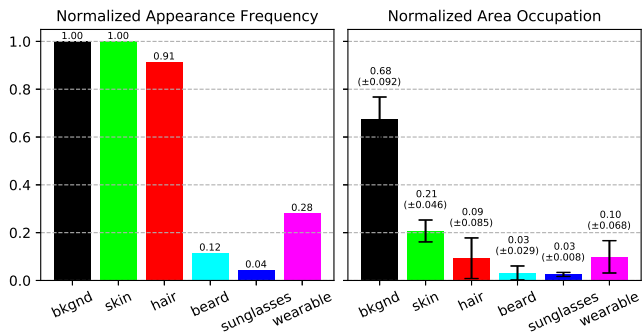}
\end{center}
\captionsep
   \caption{ELFW insights. (Left) \textit{Normalized appearance frequency} or the normalized number of class appearances per image in the whole dataset, where $1$ means that the class appears at every image. (Right) \textit{Normalized area occupation} or the proportional area occupied by each class at every image where it does appear. Note that standard deviation (top-bar vertical brackets) relates to class variability, so that, as expected, \textit{hair}, \textit{beard-moustache}, and \textit{head-wearable} are highly variable in size, while \textit{background}, \textit{skin}, and even \textit{sunglasses} have in general small variations in relation to their normalized averaged size (top-bar numbers). Furthermore, both mean and standard deviation give an idea of the maximum and minimum areas throughout the ELFW dataset, concretely occurring for \textit{background} and \textit{beard-moustache}, respectively.}
\label{fig:elfw_stats}
\end{figure}

The Extended Label Faces in-the-Wild dataset (ELFW) builds upon the LFW dataset by keeping its three original categories (\textit{background}, \textit{skin}, and \textit{hair}), extending them by relabeling cases with three additional new ones (\textit{beard-mustache}, \textit{sunglasses}, and \textit{head-wearable}), and synthetically adding facial-related objects (\textit{sunglasses}, \textit{hands}, and \textit{mouth-mask}).

The motivation under the construction of ELFW is the fact that most of the datasets with semantic annotations for face recognition do not explicitly consider objects commonly present next to faces in daily images, which can partially occlude the faces and, thus, hinder identification. As a matter of fact, LFW does not make any differences between hair and beard, sunglasses are confused by either hair or skin, hair is not properly segmented in the presence of a hat, or simply a very common object like hands occluding the face --even slightly-- is not properly handled.

For these reasons, in this section we (1) introduce ELFW, a new dataset especially constructed to deal with common facial elements and occluding objects, and (2) show means of augmenting samples with synthetic objects such as sunglasses, hands, or mouth-masks.

\subsection{Data collection}

Among the three released LFW datasets (see Sect. \ref{sec:lfw_relatedwork}), the one developed with the deep funneled approach was chosen for this work because the alignment method is publicly available and it has been reported to achieve superior face verification performance.

From the $2,927$ images annotated with the original categories (\textit{background}, \textit{skin}, and \textit{hair}), a group of $596$ was manually re-labeled from scratch because they contained at least one of the extending categories (\textit{beard-moustache}, \textit{sunglasses}, or \textit{head-wearable}). Furthermore, from the remaining not labeled images with available superpixels ---LFW was originally released with $5,749$ superpixels maps---, $827$ images having at least one of the extending categories were added up and labeled. In total, ELFW is made up of $3,754$ labeled faces, where $1,423$ have at least one of the new categories. 


\subsection{Manual ground-truth annotation methodology}
\label{sec:annotation}

The process of annotating images is never straightforward. Although difficulty varies with task, translating the simplest visual concept into a label has often multiple angles.
For instance, how to deal with teeth, are they part of the skin face or they must be left out? Do regular glasses need to be treated as sunglasses even if the eyes' contours can be seen through? To what extent does the skin along the neck need to be labeled? Are earrings a head-wearable as any regular headphones are?

On the following, we summarize the guidance instructions elaborated to annotate the dataset to have a better understanding of its labels:
\begin{itemize}
    \item For simplicity, eyes, eyebrows, mouth, and teeth are equally labeled as \textit{skin}. Neck and ears are also considered \textit{skin}.
    \item On the contrary, the skin of shoulders or hands are ignored, \ie labeled as \textit{background}.
    \item Helmets, caps, turbans, headsets, even glasses, and in general any object worn on the head ---although partially occluding the face--- lay under the same \textit{head-wearable} category.
    \item As an exception ---and because they do not generally hamper identification---, regular glasses with no color shade whatsoever are labeled as \textit{skin}.
    \item Faces at the background which do not belong to the main face are ignored (\textit{background}).
    \item Likewise, occluding objects like microphones, flags or even hands are also considered as \textit{background}.
\end{itemize}
     
Following the same annotation strategy used by LFW, the workload was alleviated by initially labeling superpixels, preserving at the same time pixel-wise accurate contours.
In order to improve productivity, a simple GUI tool was implemented to entirely annotate a superpixel with a single mouse click, having real-time visual feedback of the actual labels. In practice, scribbles were also allowed, which accelerated the annotation process. A second GUI tool allowed for manual correction of wrong segments derived from superpixels which, for instance, were outlining different categories at the same time.

The whole (re)labeling process with the new extending categories was done in $4$ weeks by $4$ different people sharing the annotating criteria  described above. The whole set of images was later manually supervised and corrected by one of the annotators. Some labeled and relabeled examples are shown in Fig.~\ref{fig:elfw_examples}.
See also Fig.~\ref{fig:elfw_stats} to get deeper insights about the contributions of each labeled category to the ELFW dataset after annotation.

\newpage
\subsection{Data augmentation}
\label{sec:data_augmentation}
Due to the high dependence on large data for training deep models, it is increasingly frequent to enlarge relatively small datasets with synthetically generated images~\cite{Gecer2018,Shrivastava2017}, which might fill the gap for real situations not depicted in the dataset and thus can help to better generalize to unseen cases, but also to balance under-represented categories. 

In this work, simple yet effective ways to automatically enlarge the proposed dataset have been used, from which ground-truth images can be trivially generated. Although these augmentations are released separately from the dataset, the code is open sourced, so that interested readers can use it at their will. The augmentation strategies are reported in the following sections.

\subsubsection{Category augmentation}

\begin{figure*}[t]
\begin{center}
\includegraphics[width=0.09\linewidth]{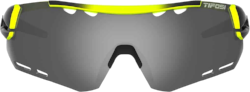}
\includegraphics[width=0.09\linewidth]{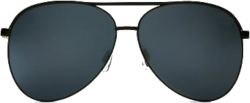}
\includegraphics[width=0.09\linewidth]{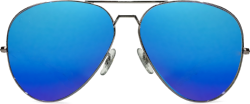}
\includegraphics[width=0.09\linewidth]{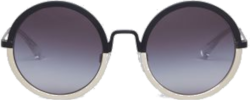}
\includegraphics[width=0.09\linewidth]{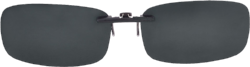}
\includegraphics[width=0.09\linewidth]{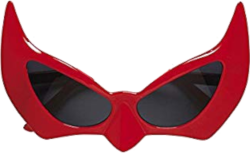}
\includegraphics[width=0.09\linewidth]{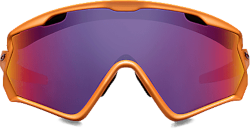}
\includegraphics[width=0.09\linewidth]{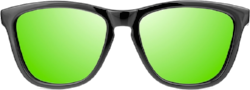}
\includegraphics[width=0.09\linewidth]{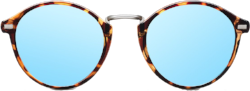}
\includegraphics[width=0.09\linewidth]{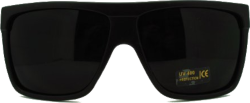}
\includegraphics[width=0.09\linewidth]{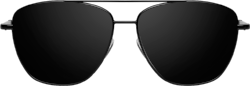}
\includegraphics[width=0.09\linewidth]{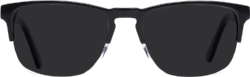}
\includegraphics[width=0.09\linewidth]{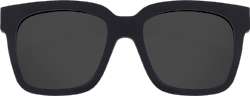}
\includegraphics[width=0.09\linewidth]{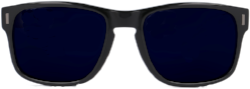}
\includegraphics[width=0.09\linewidth]{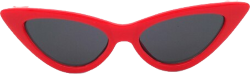}
\includegraphics[width=0.09\linewidth]{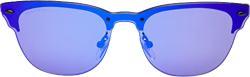}
\includegraphics[width=0.09\linewidth]{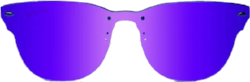}
\includegraphics[width=0.09\linewidth]{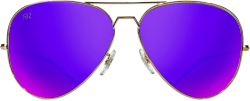}
\includegraphics[width=0.09\linewidth]{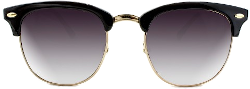}
\includegraphics[width=0.09\linewidth]{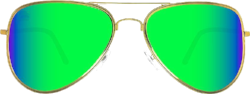}
\includegraphics[width=0.09\linewidth]{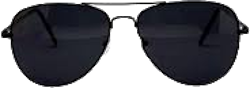}
\includegraphics[width=0.09\linewidth]{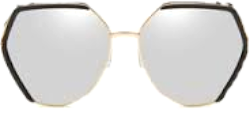}
\includegraphics[width=0.09\linewidth]{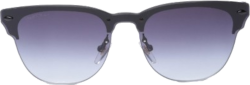}
\includegraphics[width=0.09\linewidth]{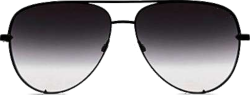}
\includegraphics[width=0.09\linewidth]{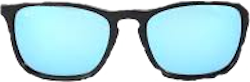}
\includegraphics[width=0.09\linewidth]{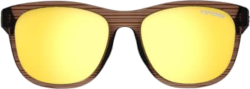}
\includegraphics[width=0.09\linewidth]{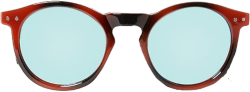}
\includegraphics[width=0.09\linewidth]{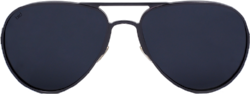}
\includegraphics[width=0.09\linewidth]{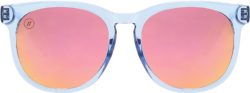}
\includegraphics[width=0.09\linewidth]{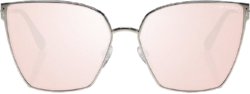}
\includegraphics[width=0.09\linewidth]{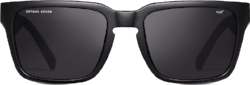}
\includegraphics[width=0.09\linewidth]{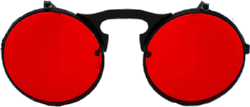}
\includegraphics[width=0.09\linewidth]{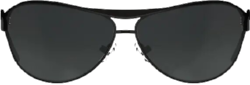}
\includegraphics[width=0.09\linewidth]{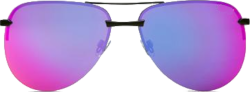}
\includegraphics[width=0.09\linewidth]{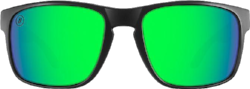}
\includegraphics[width=0.09\linewidth]{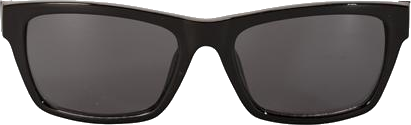}
\includegraphics[width=0.09\linewidth]{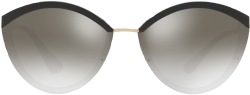}
\includegraphics[width=0.09\linewidth]{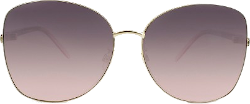}
\includegraphics[width=0.09\linewidth]{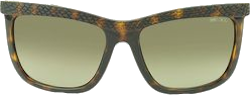}
\includegraphics[width=0.09\linewidth]{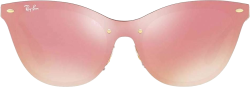}
\includegraphics[width=0.07\linewidth]{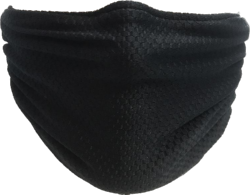}
\includegraphics[width=0.07\linewidth]{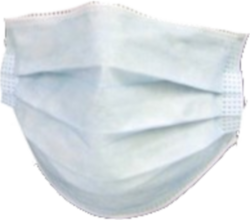}
\includegraphics[width=0.07\linewidth]{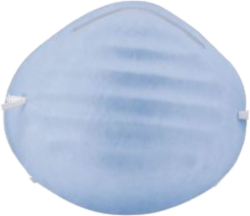}
\includegraphics[width=0.07\linewidth]{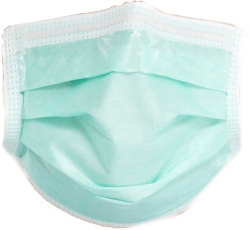}
\includegraphics[width=0.07\linewidth]{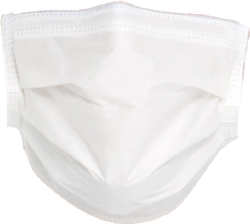}
\includegraphics[width=0.07\linewidth]{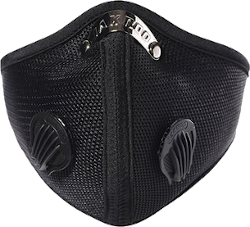}
\includegraphics[width=0.07\linewidth]{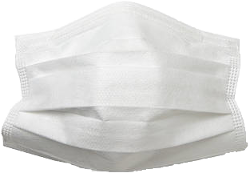}
\includegraphics[width=0.07\linewidth]{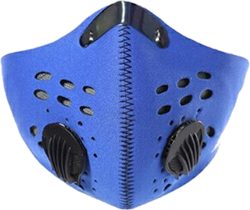}
\includegraphics[width=0.07\linewidth]{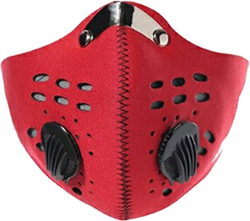}
\includegraphics[width=0.07\linewidth]{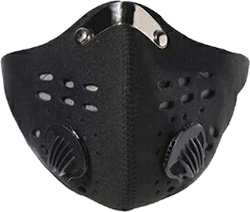}
\includegraphics[width=0.07\linewidth]{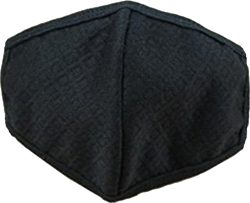}
\includegraphics[width=0.07\linewidth]{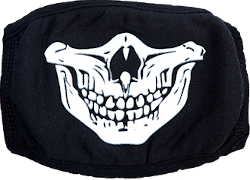}
\end{center}
\captionsep
   \caption{Augmentation assets: the $40$ sunglasses and $12$ mouth masks used for synthetic category augmentation in ELFW.}
\label{fig:augmentation_assets}
\end{figure*}

On the one hand, \textit{sunglasses} is the worst balanced category throughout the collected data, see Fig.~\ref{fig:elfw_stats}. On the other, the dataset does not present even a single case of an image with a common object typically present in faces, namely, \textit{mouth masks}. When present, both types of objects usually occlude a large facial area and impede identification. In their turn, though, it is particularly easy to automatically add them to a given face.
To this end, $40$ diverse types of sunglasses and $12$ diverse types of mouth masks were obtained from the Internet and manually retouched to guarantee an appropriate blending processing with a face. The whole collection of augmentation assets is depicted in Fig.~\ref{fig:augmentation_assets}.

\begin{figure}[t]
\begin{center}
\includegraphics[width=0.099\linewidth]{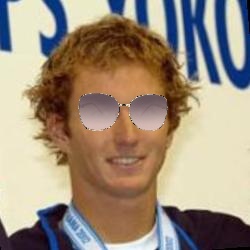}\figcolsep
\includegraphics[width=0.099\linewidth]{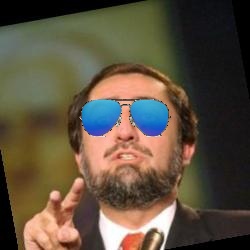}\figcolsep
\includegraphics[width=0.099\linewidth]{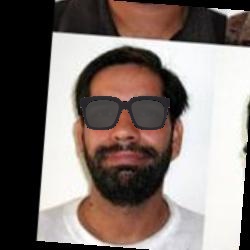}\figcolsep
\includegraphics[width=0.099\linewidth]{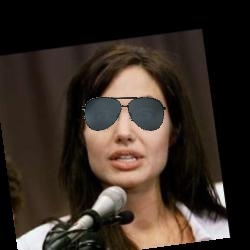}\figcolsep
\includegraphics[width=0.099\linewidth]{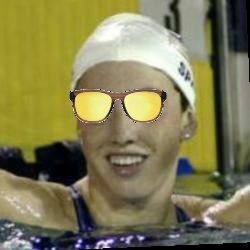}\figcolsep
\includegraphics[width=0.099\linewidth]{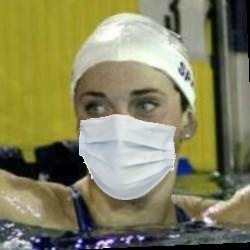}\figcolsep
\includegraphics[width=0.099\linewidth]{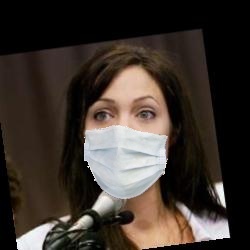}\figcolsep
\includegraphics[width=0.099\linewidth]{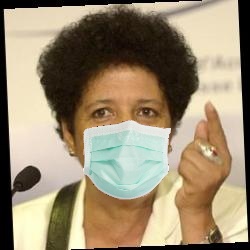}\figcolsep
\includegraphics[width=0.099\linewidth]{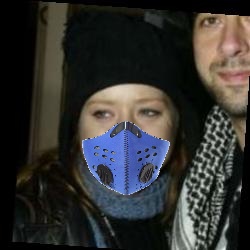}\figcolsep
\includegraphics[width=0.099\linewidth]{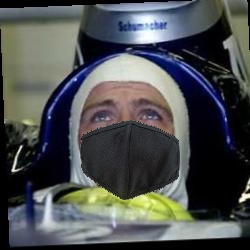}\figcolsep

\figrowsep

\includegraphics[width=0.099\linewidth]{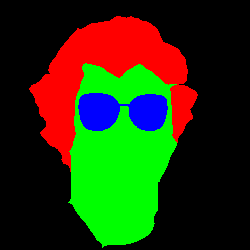}\figcolsep
\includegraphics[width=0.099\linewidth]{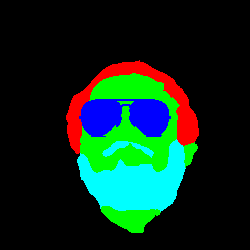}\figcolsep
\includegraphics[width=0.099\linewidth]{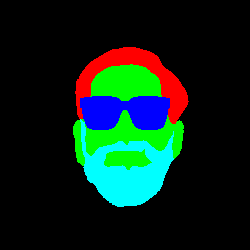}\figcolsep
\includegraphics[width=0.099\linewidth]{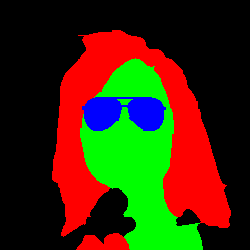}\figcolsep
\includegraphics[width=0.099\linewidth]{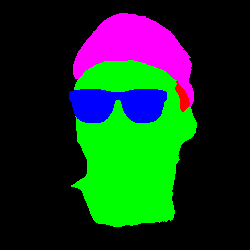}\figcolsep
\includegraphics[width=0.099\linewidth]{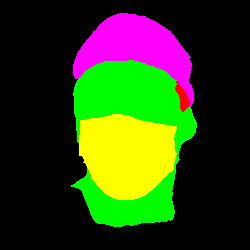}\figcolsep
\includegraphics[width=0.099\linewidth]{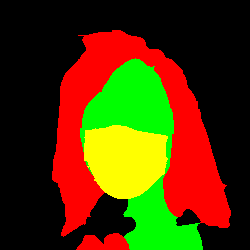}\figcolsep
\includegraphics[width=0.099\linewidth]{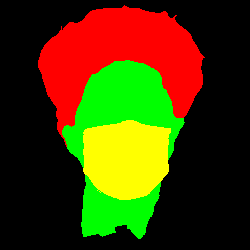}\figcolsep
\includegraphics[width=0.099\linewidth]{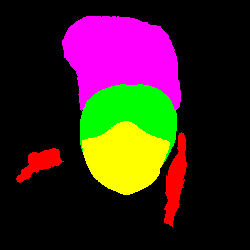}\figcolsep
\includegraphics[width=0.099\linewidth]{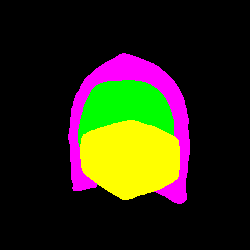}\figcolsep

\end{center}
\captionsep
   \caption{Examples of automatic category augmentation with \textit{sunglasses} and \textit{mouth-mask} in ELFW.}
\label{fig:elfw_augmentation}
\end{figure}

By construction, all LFW faces should be detected in principle by the Viola-Jones algorithm~\cite{viola2004robust}, but interestingly not all of them were so detected by the used OpenCV implementation. Then, a total of $2,003$ faces were suitable for category augmentation. 
Before being attached, the sunglasses assets were resized proportionally to the interocular distance and made slightly transparent. Similarly, this distance was also used as a reference to estimate the face size to properly resize the masks. Note that no shading nor color correction was applied here, and although an artificial appearance is patent in some cases, 
the generated cases were effective enough to reinforce the category learning stage in a semantic segmentation scenario, see Sec.~\ref{sec:experimental_results}. Examples of the described category augmentations can be seen in Fig.~\ref{fig:elfw_augmentation}.


\subsubsection{Augmenting by occluding}

There are other common occluding objects in natural images but not attached to the face itself. As a matter of fact, LFW contains an important amount of them, such as a multiplicity of hand-held objects like microphones, rackets, sport balls, or even other faces. In fact, the variety of occlusions can be as broad as the variety of conditions for acquiring `clean' faces.
In particular, hands are, by nature, one of the most frequent elements among occluders~\cite{mahmoud2011interpreting}. It is, however, an specially challenging \textit{object} since the skin color shared with the face can entangle posterior face-hand discrimination. For these reasons, it appears reasonable to additionally use hands for data augmentation.

Blending source hands into faces requires realistic color and pose matching with respect to the targeted face. 
To the extent of our knowledge, the most determined work on this regard is Hand2Face~\cite{nojavanasghari2017hand2face}, whose authors gave especial relevance to hands because their pose discloses relevant information about the person's affective state. 
Other hand datasets made of images captured in first person view, such as Egohands~\cite{bambach2015lending}, GTEA~\cite{li2015delving}, or EgoYouTubeHands~\cite{urooj2018analysis} are not suitable to be attached to faces in a natural way. The latter work, however, also unveiled a significant hand dataset in third person view, HandOverFace~\cite{urooj2018analysis}. In the end, the hands used in this work were compiled from both Hand2Face and HandOverFace.



\begin{figure*}[t]
\begin{center}

\includegraphics[width=0.099\linewidth]{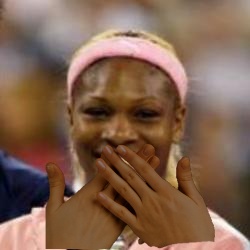}\figcolsep
\includegraphics[width=0.099\linewidth]{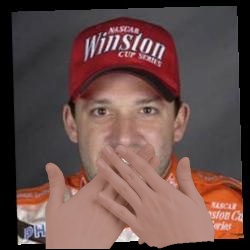}\figcolsep
\includegraphics[width=0.099\linewidth]{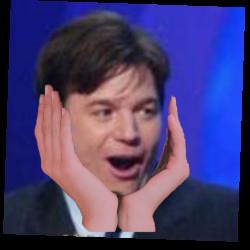}\figcolsep
\includegraphics[width=0.099\linewidth]{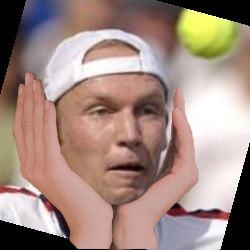}\figcolsep
\includegraphics[width=0.099\linewidth]{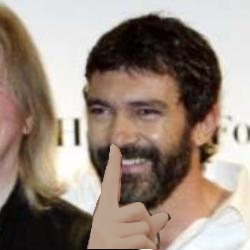}\figcolsep
\includegraphics[width=0.099\linewidth]{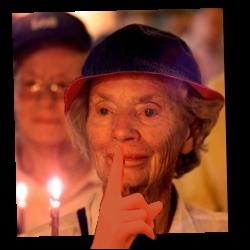}\figcolsep
\includegraphics[width=0.099\linewidth]{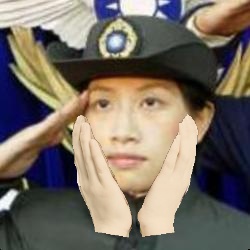}\figcolsep
\includegraphics[width=0.099\linewidth]{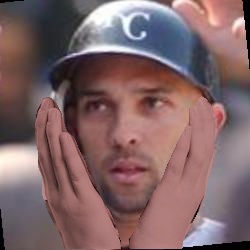}\figcolsep
\includegraphics[width=0.099\linewidth]{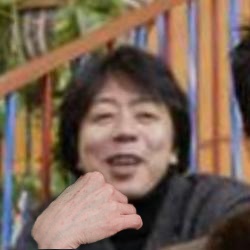}\figcolsep
\includegraphics[width=0.099\linewidth]{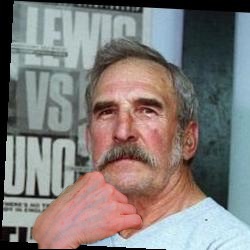}\figcolsep

\figrowsep

\includegraphics[width=0.099\linewidth]{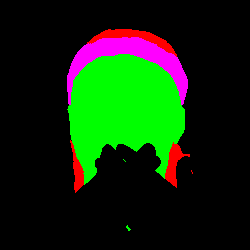}\figcolsep
\includegraphics[width=0.099\linewidth]{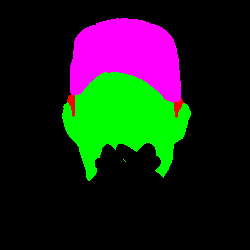}\figcolsep
\includegraphics[width=0.099\linewidth]{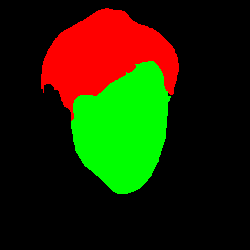}\figcolsep
\includegraphics[width=0.099\linewidth]{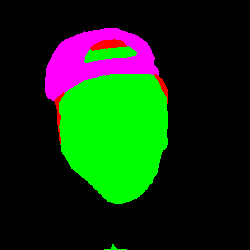}\figcolsep
\includegraphics[width=0.099\linewidth]{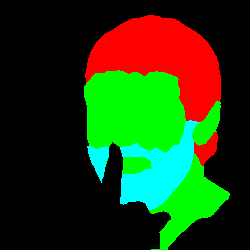}\figcolsep
\includegraphics[width=0.099\linewidth]{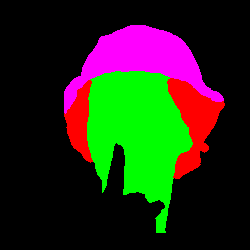}\figcolsep
\includegraphics[width=0.099\linewidth]{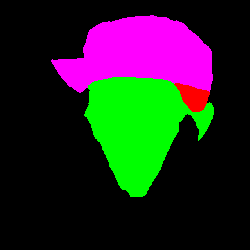}\figcolsep
\includegraphics[width=0.099\linewidth]{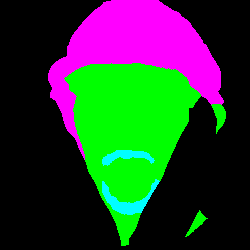}\figcolsep
\includegraphics[width=0.099\linewidth]{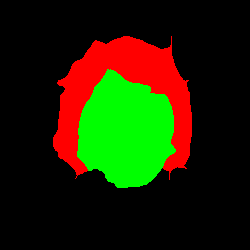}\figcolsep
\includegraphics[width=0.099\linewidth]{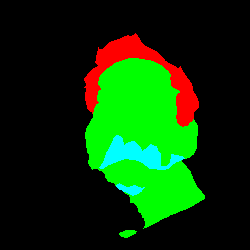}\figcolsep

\figrowsep

\includegraphics[width=0.099\linewidth]{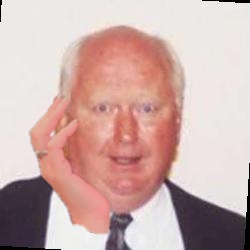}\figcolsep
\includegraphics[width=0.099\linewidth]{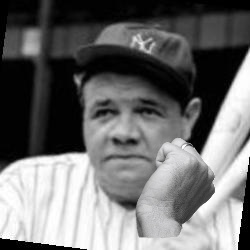}\figcolsep
\includegraphics[width=0.099\linewidth]{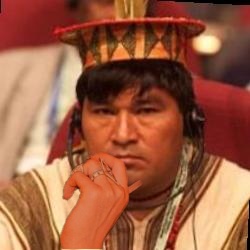}\figcolsep
\includegraphics[width=0.099\linewidth]{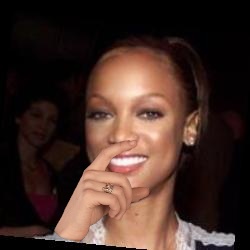}\figcolsep
\includegraphics[width=0.099\linewidth]{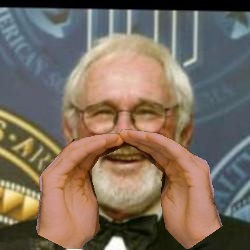}\figcolsep
\includegraphics[width=0.099\linewidth]{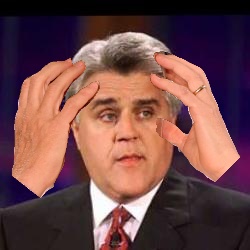}\figcolsep
\includegraphics[width=0.099\linewidth]{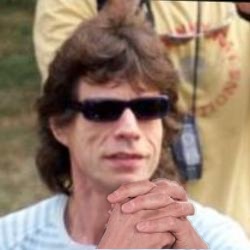}\figcolsep
\includegraphics[width=0.099\linewidth]{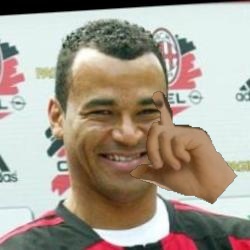}\figcolsep
\includegraphics[width=0.099\linewidth]{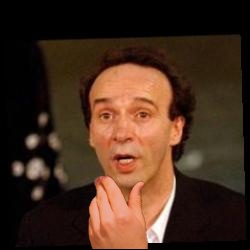}\figcolsep
\includegraphics[width=0.099\linewidth]{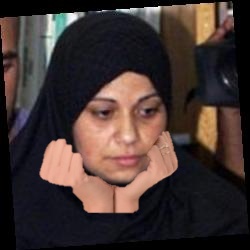}\figcolsep

\figrowsep

\includegraphics[width=0.099\linewidth]{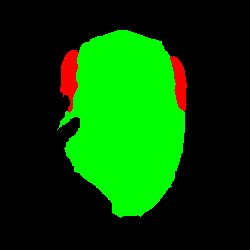}\figcolsep
\includegraphics[width=0.099\linewidth]{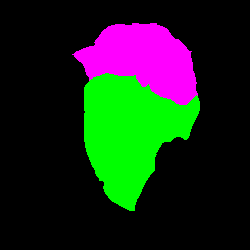}\figcolsep
\includegraphics[width=0.099\linewidth]{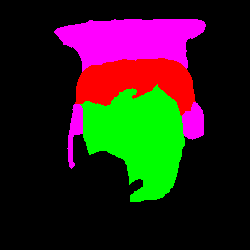}\figcolsep
\includegraphics[width=0.099\linewidth]{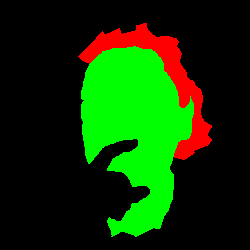}\figcolsep
\includegraphics[width=0.099\linewidth]{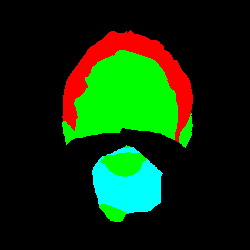}\figcolsep
\includegraphics[width=0.099\linewidth]{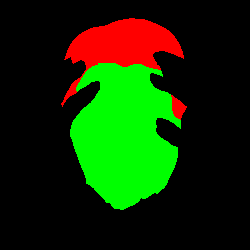}\figcolsep
\includegraphics[width=0.099\linewidth]{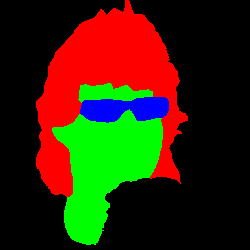}\figcolsep
\includegraphics[width=0.099\linewidth]{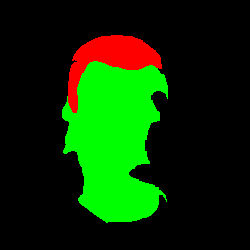}\figcolsep
\includegraphics[width=0.099\linewidth]{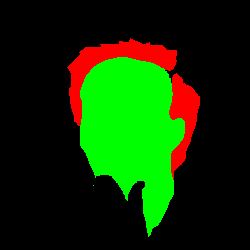}\figcolsep
\includegraphics[width=0.099\linewidth]{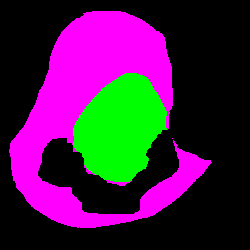}\figcolsep

\end{center}
\captionsep
\caption{Illustration of occluding hands synthetically attached to ELFW. The first and second rows (images and labels) show paired examples of re-used hands on different faces with their corresponding automatic color tone correction, scaling, and positioning. The third and fourth rows (images and labels) depict an assortment of hand poses, with the same automatic attachment procedure.}
\vspace{-10pt}
\label{fig:elfw_occluding}
\end{figure*}

\begin{figure}[t]
\begin{center}
	\includegraphics[width=1.0\linewidth]{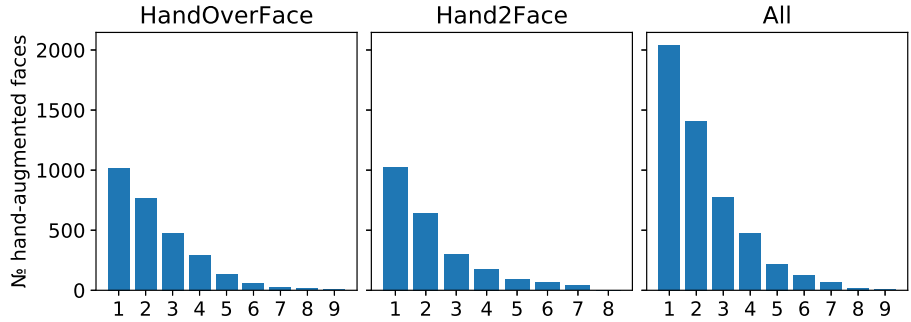}
\end{center}
\captionsep
   \caption{Histograms illustrating the number of faces (vertical axis) which have been augmented with an specific number of different hands (horizontal axis). Such distributions are shaped by using $\theta=5^\circ$ head pose match between the source faces in HandOverFace or Hand2Face and the target faces in ELFW. It all adds up to $11,572$ new cases out of $3,153$ different faces and $177$ different hands.}
\label{fig:elfw_stats_augmentation}
\end{figure}

To attach hands to faces we basically followed the approach described in~\cite{nojavanasghari2017hand2face}. Firstly, the source head pose ---originally with hands on it--- is matched against all the target head poses in ELFW by using Dlib~\cite{king2009dlib}. Two poses match if the distance between them ---measured as the $L_2$ norm of the elevation, azimuth and rotation angles--- is under a given solid angle threshold $\theta$. Secondly, the hands color is corrected according to each facial tone. For that, the averaged face color is measured inside a rectangular area containing the nose and both cheeks, which is largely uncovered for most of the current faces. The averaged color is transferred from the target face to the source hands by correcting the mean and standard deviation for each channel in the $l\alpha\beta$ color space~\cite{reinhard2001color}. Before being attached, as with category augmentation, the hands are resized by using a scale factor relative to both origin and destination face sizes. Likewise, hands are also centered by considering their relative location from the source face center to the destination face center. Multiple examples of synthetic occluding hands are showed in Fig.~\ref{fig:elfw_occluding} and the final hands usage distribution is illustrated in Fig.~\ref{fig:elfw_stats_augmentation}.

\section{Experiments: facial semantic segmentation}

In this section we provide benchmarking results on the ELFW dataset for semantic segmentation, a natural evaluation framework for this released type of data. We chose two deep neural networks in the semantic segmentation state-of-the-art, namely the Fully Convolutional Network (FCN) originally proposed in the seminal work~\cite{long2015fully}, and a much recent architecture such as DeepLabV3~\cite{chen2017deeplab}, which has been reported to perform remarkably well in several standard datasets.

\subsection{Baseline configuration}
\label{sec:baseline_config}

All experiments carried out here were configured alike in order to provide a common baseline for assessing the ELFW dataset.

\medskip\noindent\textbf{Training hyper-parameters}: Both models, FCN and DeepLabV3, used a COCO train2017 pre-trained ResNet-101 backbone. Then, they were fine-tuned by minimizing a pixel-wise cross entropy loss under a $16$ batch sized SGD optimizer, and by scheduling a multi-step learning rate to provoke an initial fast learning with $10^{-3}$ at the very first epochs, latter lowered to $2^{-4}$ at epoch $35$, and finally $0.4^{-5}$ at epoch $90$, where the performance stabilized. Weight decay was set to $5^{-4}$ and momentum to $0.99$. In all experiments basic data augmentation such as random horizontal flips, affine shifts and image resize transformations were performed. An early-stop was set to 30 epochs without improving.

\medskip\noindent\textbf{Augmentation factor}:
In order to evaluate the proposed category augmentation strategies we defined $\sigma$ to be the factor denoting the number of object-based augmentation faces added to the main non-augmented training dataset. For instance, $\sigma=0.5$ means that an extra $50\%$ with respect to the base train set ($3,754$) of synthetically augmented images have been added for training, \ie $1,877$ augmented images. When more than one augmentation asset is used, this factor is uniformly distributed among the different augmentation types.

\medskip\noindent\textbf{Validation sets}:
To assess each of the augmented categories, separately and jointly, we used the same validation being careful with the nature of each data type. It was defined as $10\%$ of the base train set, \ie $376$ images in total, where $62$ out of $125$ images wore real sunglasses and $28$ out of $56$ faces were originally occluded by real hands. The remaining halves were reserved for training. The validation set was then randomly populated with other faces from the remaining set.

\medskip\noindent\textbf{Hardware and software}: 
Each of the neural models was trained on an individual Nvidia GTX 1080Ti GPU. Regardless of the network architecture, the configuration with the smallest dataset ($\sigma=0$) required about $1$ training day, while the largest one ($\sigma=1$) took about a week.
We employed the frameworks PyTorch $1.1.0$ and TorchVision $0.3.0$ versions, which officially released implementations of both FCN and DeeplabV3 models with the ResNet-101 pre-trained backbones\footnote{\url{ https://pytorch.org/docs/stable/torchvision/models.html}.}.

\subsection{Validation tests}
\label{sec:experimental_results}

\begin{table}[t]
    \centering
        \caption{\textbf{Global metrics:} results for both FCN and DeepLabV3 architectures with different augmentation assets (sunglasses, hands and both) and an increasing augmentation factor ($\sigma$). The best result is highlighted in bold for each quality metric. The performance gain indicates the difference between the highest value among $\sigma>0$ and the baseline $\sigma=0$. The same validation set was used for all experiments.
        }
    \label{tab:global_results}
    \resizebox{1.0\textwidth}{!}
    {
    \begin{tabular}{cccccccccccccccccc}
    \toprule
    && &  &\multicolumn{4}{c}{\textbf{Sunglasses augmentation}} &
    &\multicolumn{4}{c}{\textbf{Hands augmentation}} &
    &\multicolumn{4}{c}{\textbf{Both types (sunglasses + hands)}}\\
    \cmidrule{5-8} \cmidrule{10-13} \cmidrule{15-18}
    && & & Pixel Acc. & Mean Acc.  & Mean IU    & Freq.W. IU 
      & & Pixel Acc. & Mean Acc.  & Mean IU    & Freq.W. IU 
      & & Pixel Acc. & Mean Acc.  & Mean IU    & Freq.W. IU   \\
    \midrule
    \parbox[t]{2mm}{\multirow{5}{*}{\rotatebox[origin=c]{90}{\normalsize FCN}}}
&& $\sigma=0.00$ &&$94.86$	&$\textbf{89.95}$	&$78.54$	&$90.62$		&&$94.86$	&$89.95$	&$78.54$	&$90.62$		&&$94.86$	&$89.95$	&$78.54$	&$90.62$ \\
&&\cellcolor{blue!8}  $\sigma=0.25$ &&$94.88$	&$89.35$	&$78.57$	&$90.65$		&&$94.92$	&$\textbf{90.04}$	&$78.89$	&$90.72$		&&$94.87$	&$90.01$	&$79.12$	&$90.65$ \\
&&\cellcolor{blue!16} $\sigma=0.50$ &&$94.82$	&$89.07$	&$78.58$	&$90.55$		&&$\textbf{95.01}$	&$89.30$	&$\textbf{79.28}$	&$\textbf{90.86}$		&&$\textbf{94.92}$	&$89.82$	&$\textbf{79.52}$	&$\textbf{90.71}$ \\
&&\cellcolor{blue!22} $\sigma=1.00$ &&$\textbf{94.91}$	&$88.96$	&$\textbf{79.15}$	&$\textbf{90.68}$		&&$94.94$	&$89.36$	&$79.08$	&$90.75$		&&$94.90$	&$\textbf{90.06}$	&$79.42$	&$90.70$ \\
\cmidrule{2-18}
&& \cellcolor{red!10} Gain 		   &&$0.05$		&$-0.60$	&$0.61$		&$0.06$			&&$0.15$	&$0.09$		&$0.74$		&$0.24$			&&$0.06$	&$0.11$		&$0.98$		&$0.09$ \\
    \midrule
    \parbox[t]{2mm}{\multirow{5}{*}{\rotatebox[origin=c]{90}{\normalsize DeepLabV3}}}
&& $\sigma=0.00$ &&$94.68$	&$89.71$	&$77.95$	&$90.37$		&&$94.68$	&$89.71$	&$77.95$	&$90.37$		&&$94.68$	&$89.71$	&$77.95$	&$90.37$ \\
&&\cellcolor{blue!8}  $\sigma=0.25$ &&$94.79$	&$89.60$	&$78.57$	&$90.51$		&&$94.87$	&$89.75$	&$78.62$	&$90.63$		&&$94.81$	&$89.92$	&$78.41$	&$90.56$ \\
&&\cellcolor{blue!16} $\sigma=0.50$ &&$94.83$	&$\textbf{90.35}$	&$79.05$	&$90.58$		&&$94.86$	&$\textbf{90.18}$	&$78.72$	&$90.64$		&&$\textbf{94.90}$	&$90.12$	&$79.14$	&$\textbf{90.71}$ \\
&&\cellcolor{blue!22} $\sigma=1.00$ &&$\textbf{94.89}$	&$90.07$	&$\textbf{79.39}$	&$\textbf{90.68}$		&&$\textbf{94.94}$	&$90.08$	&$\textbf{78.88}$	&$\textbf{90.75}$		&&$\textbf{94.90}$	&$\textbf{90.38}$	&$\textbf{79.38}$	&$90.70$ \\
\cmidrule{2-17}
&& \cellcolor{red!10} Gain 		   &&$0.21$	    &$0.64$	    &$1.44$	    &$0.31$		    &&$0.26$	&$0.47$	    &$0.93$	    &$0.38$		    &&$0.22$	&$0.67$	    &$1.43$	    &$0.34$ \\
    \bottomrule
    \end{tabular}
    }

\end{table}

\begin{table}[t]
    \centering
        \caption{\textbf{Class IU:} class-wise intersection over union performance for both FCN and DeepLabV3 augmented with sunglasses, hands, and both according to the augmentation factor ($\sigma$). Details shared alike with Tab.~\ref{tab:global_results}.
        }
    \label{tab:classwise_results}
    \resizebox{1.0\textwidth}{!}
    {
    \begin{tabular}{cccccccccccccccccccccccc}
    \toprule
    && &  &\multicolumn{6}{c}{\textbf{Sunglasses augmentation}} &
    &\multicolumn{6}{c}{\textbf{Hands augmentation}} &
    &\multicolumn{6}{c}{\textbf{Both types (sunglasses + hands)}}\\
    \cmidrule{5-10} \cmidrule{12-17} \cmidrule{19-24}
    && & & bkgnd & skin  & hair    & beard   & snglss & wear 
      & & bkgnd & skin  & hair    & beard   & snglss & wear
      & & bkgnd & skin  & hair    & beard   & snglss & wear \\
    \midrule
    \parbox[t]{2mm}{\multirow{5}{*}{\rotatebox[origin=c]{90}{\normalsize FCN}}}
&&  $\sigma=0.00$ &&$\textbf{94.76}$	&$86.38$	&$71.86$	&$61.34$	&$72.45$	&$84.47$		&&$94.76$	&$86.38$	&$71.86$	&$61.34$	&$72.45$	&$84.47$		&&$94.76$	&$86.38$	&$\textbf{71.86}$	&$61.34$	&$72.45$	&$84.47$ \\
&&\cellcolor{blue!8}  $\sigma=0.25$ &&$94.75$	&$86.53$	&$\textbf{71.93}$	&$60.09$	&$73.82$	&$84.34$		&&$94.78$	&$86.77$	&$71.78$	&$61.24$	&$74.51$	&$84.26$		&&$94.74$	&$86.72$	&$71.41$	&$63.02$	&$74.86$	&$83.97$ \\
&&\cellcolor{blue!16} $\sigma=0.50$ &&$94.71$	&$86.49$	&$71.23$	&$60.27$	&$74.76$	&$84.02$		&&$\textbf{94.92}$	&$\textbf{86.89}$	&$\textbf{72.04}$	&$\textbf{62.55}$	&$\textbf{75.33}$	&$83.95$		&&$\textbf{94.79}$	&$86.70$	&$71.53$	&$\textbf{63.19}$	&$\textbf{75.80}$	&$\textbf{85.11}$ \\
&&\cellcolor{blue!22} $\sigma=1.00$ &&$94.75$	&$\textbf{86.68}$	&$71.69$	&$\textbf{62.33}$	&$\textbf{74.78}$	&$\textbf{84.64}$		&&$94.80$	&$86.70$	&$71.98$	&$61.76$	&$73.98$	&$\textbf{85.25}$		&&$94.73$	&$\textbf{86.73}$	&$71.77$	&$62.77$	&$75.48$	&$85.04$ \\
\cmidrule{2-24}
&& \cellcolor{red!10} Gain 		   &&$-0.01$	&$0.30$		&$0.07$		&$0.99$		&$2.33$		&$0.17$			&&$0.16$	&$0.51$		&$0.18$		&$1.21$		&$2.88$		&$0.78$			&&$0.03$	&$0.35$		&$-0.09$	&$1.85$		&$3.35$		&$0.64$ \\
    \midrule
    \parbox[t]{2mm}{\multirow{5}{*}{\rotatebox[origin=c]{90}{\normalsize DeepLabV3}}}
&& $\sigma=0.00$ &&$94.51$	&$86.31$	&$71.33$	&$60.57$	&$71.35$	&$83.62$		&&$94.51$	&$86.31$	&$71.33$	&$60.57$	&$71.35$	&$83.62$		&&$94.51$	&$86.31$	&$71.33$	&$60.57$	&$71.35$	&$83.62$ \\
&&\cellcolor{blue!8}  $\sigma=0.25$ &&$94.64$	&$86.42$	&$71.66$	&$61.14$	&$74.20$	&$83.36$		&&$94.77$	&$86.44$	&$71.85$	&$62.07$	&$72.95$	&$83.65$		&&$94.66$	&$86.51$	&$71.66$	&$59.54$	&$74.23$	&$83.86$ \\
&&\cellcolor{blue!16} $\sigma=0.50$ &&$94.67$	&$86.37$	&$\textbf{71.90}$	&$61.60$	&$75.30$	&$84.48$		&&$94.73$	&$86.60$	&$71.78$	&$\textbf{62.61}$	&$72.75$	&$83.85$		&&$\textbf{94.76}$	&$86.66$	&$\textbf{71.85}$	&$61.45$	&$74.97$	&$\textbf{85.17}$ \\
&&\cellcolor{blue!22} $\sigma=1.00$ &&$\textbf{94.75}$	&$\textbf{86.63}$	&$71.73$	&$\textbf{63.27}$	&$\textbf{75.43}$	&$\textbf{84.52}$		&&$\textbf{94.81}$	&$\textbf{86.64}$	&$\textbf{72.29}$	&$61.98$	&$\textbf{73.22}$	&$\textbf{84.31}$		&&$94.72$	&$\textbf{86.84}$	&$71.83$	&$\textbf{63.31}$	&$\textbf{75.26}$	&$84.30$ \\
\cmidrule{2-24}																			
&& \cellcolor{red!10} Gain 		   &&$0.24$	&$0.32$		&$0.57$		&$2.70$		&$4.08$		&$0.90$			&&$0.30$	&$0.33$		&$0.96$		&$2.04$		&$1.87$		&$0.69$			&&$0.25$	&$0.53$		&$0.52$		&$2.74$		&$3.91$		&$1.55$ \\
    \bottomrule
    \end{tabular}
    }

\end{table}

\begin{figure}[t]
\begin{center}
	\includegraphics[width=0.82\linewidth]{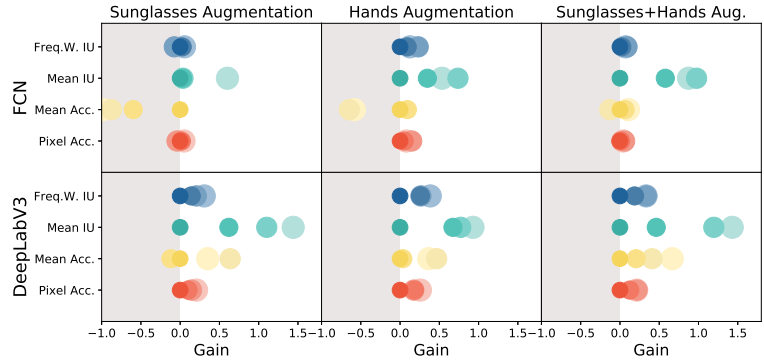}
\end{center}
\captionsep
   \caption{Gain effect with different data augmentation types and ratios ($\sigma$) on global metrics for both FCN and DeepLabV3 architectures. The size of each training dataset (related to $\sigma$) is proportionally represented by each circular area.}
\label{fig:elfw_results_metrics}
\end{figure}

\begin{figure}[h]
\begin{center}
	\includegraphics[width=0.8\linewidth]{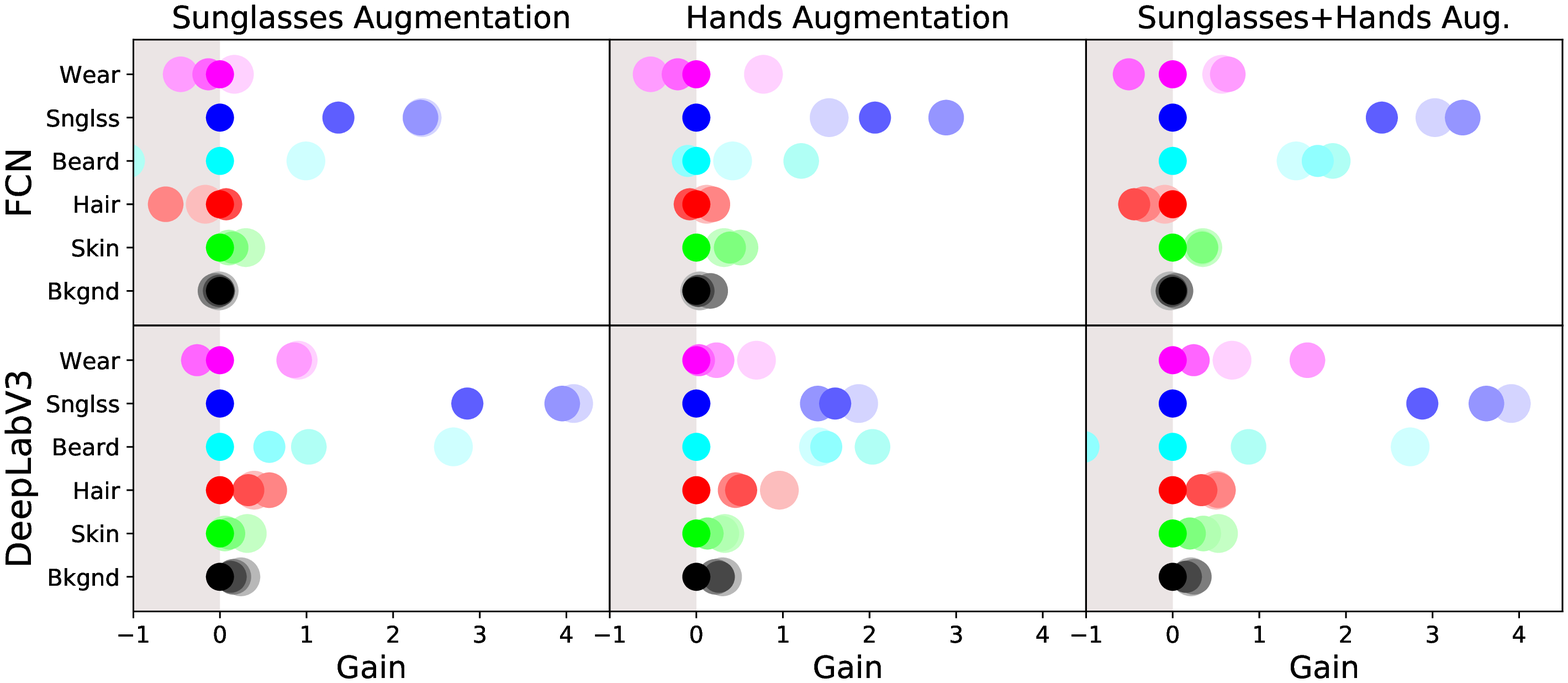}
\end{center}
\captionsep
   \caption{Gain effect per class on Mean IU with different data augmentation types and ratios ($\sigma$) for both FCN and DeepLabV3 architectures. The size of each training dataset (related to $\sigma$) is proportionally represented by each circular area.}
\label{fig:elfw_results_classes}
\end{figure}

To report results, the same metrics in~\cite{long2015fully} are considered, namely \textit{Pixel Accuracy}, \textit{Mean Accuracy}, \textit{Mean IU} and \textit{Frequency Weighted IU}. Pixel Accuracy is a class-independent global measure determining the ratio of correctly classified pixels. Mean Accuracy averages the corresponding true positive ratio across classes. Mean IU averages the intersection over union across classes. While Frequency Weighted IU weights the IU values by the corresponding normalized class appearances.
In Tab.~\ref{tab:global_results} both FCN and DeepLabV3 performances on the four global metrics are shown for sunglasses, hands, and both assets-based augmentation types and increasing factors.
For each architecture and augmentation type, the reported values on all metrics correspond to the epoch ---thus the same trained model--- where the Mean IU reaches a maximum along the whole training stage. Mean IU is chosen over the others since it considers more accuracy factors and equally balanced classes. A visual representation of the performance gain is presented in Fig.~\ref{fig:elfw_results_metrics}.

Overall, both augmentation techniques improve, although not always steady, the segmentation accuracy for both networks. Indeed, $\sigma=1$ tends to deliver the best results, \ie the more augmentation data, generally the better.
DeepLabV3, performed on pair with FCN along the experiments. Although DeeplabV3 provides interesting multi-scale features, they probably do not make much of a difference in a dataset like this, since segments actually mostly preserve their size across images.


Since all metrics are global, it is somehow hidden which of the classes are improving or deteriorating. However, Mean IU ---which averages across all classes--- showed a higher gain, revealing that some classes are certainly being boosted.
In Tab.~\ref{tab:classwise_results}, the scores are dissected per class, in which Class IU is taken over Class Accuracy because the former supplements with a false positives factor.
Reported values correspond to the exact same  models shown in Tab.~\ref{tab:global_results}. In Fig.~\ref{fig:elfw_results_classes} the gain per class versus augmentation data is illustrated.
While FCN behaved irregularly, DeepLabV3 was able to take more profit from the larger augmented instances of the training set. Moreover, \textit{sunglasses} experienced a higher gain when augmentation considered only sunglasses, \textit{beard} was the one for hands augmentation ---an expected outcome since hands tend to occlude beards---, and the exact same two categories underwent the highest boost when both augmentation categories were used. 



\subsection{Field experiments}

In this section, segmentation is qualitatively evaluated under different laboratory situations. In particular, we want to visualize the models' generalization capacity on the \textit{mouth-mask} category, which was purely synthetically added to the training set. 
The deployed model was an FCN trained with the three augmentation categories for $\sigma = 0.5$ at the best checkpoint for the global Mean IU across all epochs on the validation set described in Sec.~\ref{sec:baseline_config}.

\begin{figure}[t]

\begin{center}
\includegraphics[width=0.142\linewidth]{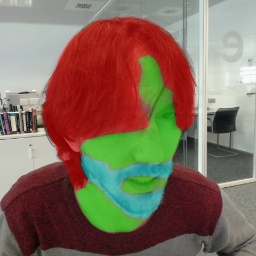}\figcolsep
\includegraphics[width=0.142\linewidth]{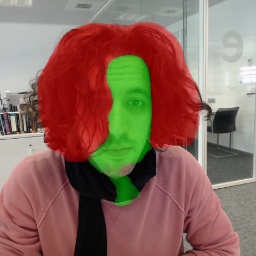}\figcolsep
\includegraphics[width=0.142\linewidth]{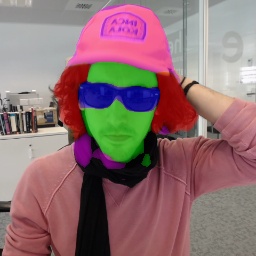}\figcolsep
\includegraphics[width=0.142\linewidth]{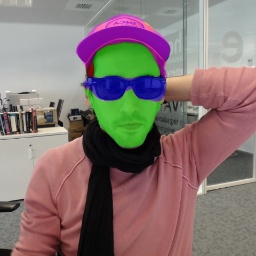}\figcolsep
\includegraphics[width=0.142\linewidth]{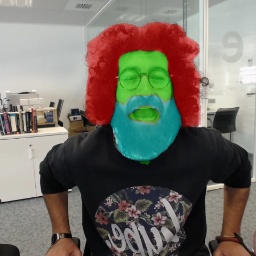}\figcolsep
\includegraphics[width=0.142\linewidth]{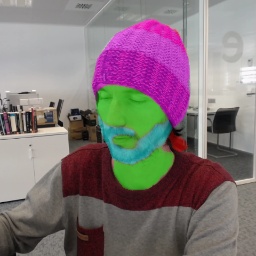}\figcolsep
\includegraphics[width=0.142\linewidth]{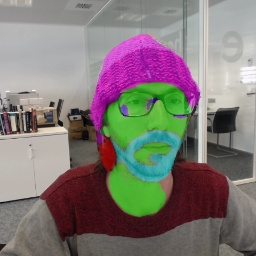}\figcolsep

\figrowsep

\includegraphics[width=0.142\linewidth]{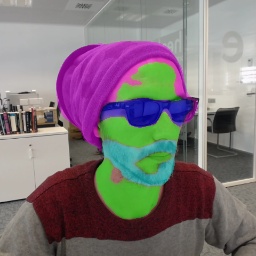}\figcolsep
\includegraphics[width=0.142\linewidth]{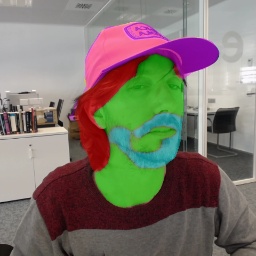}\figcolsep
\includegraphics[width=0.142\linewidth]{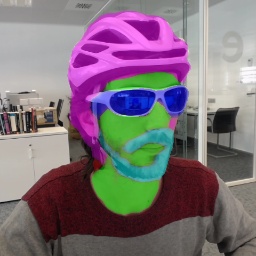}\figcolsep
\includegraphics[width=0.142\linewidth]{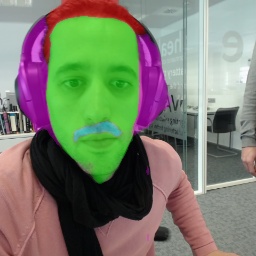}\figcolsep
\includegraphics[width=0.142\linewidth]{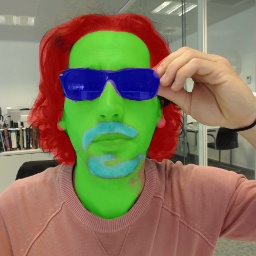}\figcolsep
\includegraphics[width=0.142\linewidth]{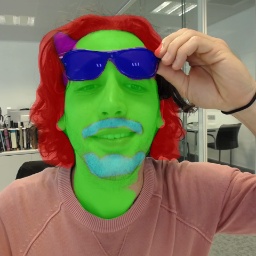}\figcolsep
\includegraphics[width=0.142\linewidth]{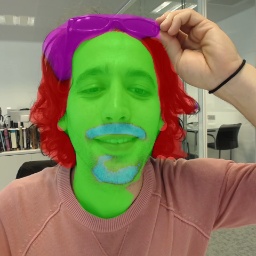}\figcolsep

\end{center}
\captionsep
   \caption{Labeling faces at the lab: sample assortment of successful detections of a casual variety of objects and poses never seen before by the neural network. }
\label{fig:webcam_success}
\end{figure}

\begin{figure}[t]

\begin{center}
\includegraphics[width=0.245\linewidth]{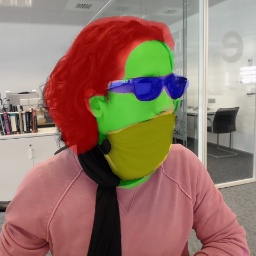}\figcolsep
\includegraphics[width=0.245\linewidth]{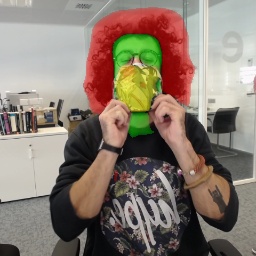}\figcolsep
\includegraphics[width=0.245\linewidth]{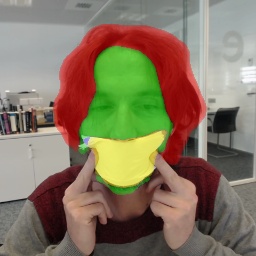}\figcolsep
\includegraphics[width=0.245\linewidth]{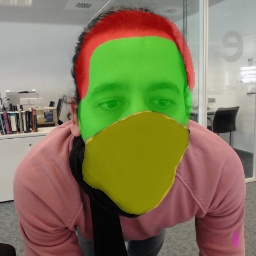}\figcolsep
\end{center}
\captionsep
   \caption{Labeling faces at the lab: \textit{mouth-masks} detection. Even though \textit{mouth-mask} only occurs synthetically in the train set, the segmentation network is able to generally give accurate results for such category.}
\label{fig:webcam_mouthmasks}
\end{figure}


\begin{figure}[t]
\begin{center}
\includegraphics[width=0.195\linewidth]{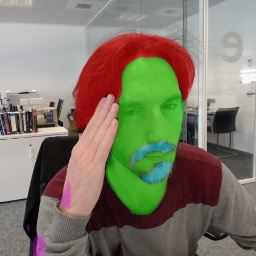}\figcolsep
\includegraphics[width=0.195\linewidth]{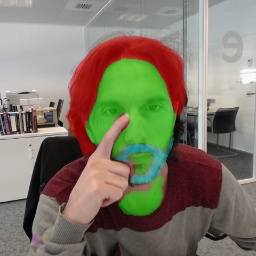}\figcolsep
\includegraphics[width=0.195\linewidth]{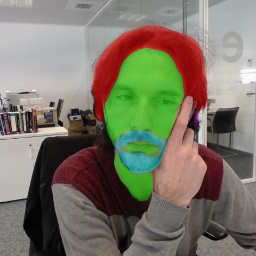}\figcolsep
\includegraphics[width=0.195\linewidth]{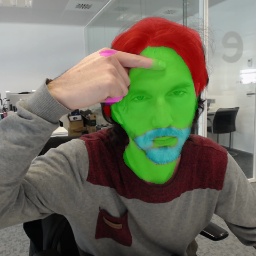}\figcolsep
\includegraphics[width=0.195\linewidth]{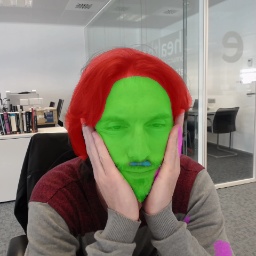}\figcolsep

\figrowsep

\includegraphics[width=0.195\linewidth]{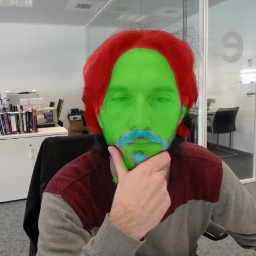}\figcolsep
\includegraphics[width=0.195\linewidth]{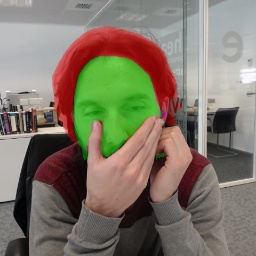}\figcolsep
\includegraphics[width=0.195\linewidth]{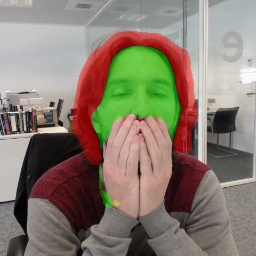}\figcolsep
\includegraphics[width=0.195\linewidth]{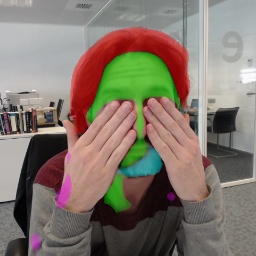}\figcolsep
\includegraphics[width=0.195\linewidth]{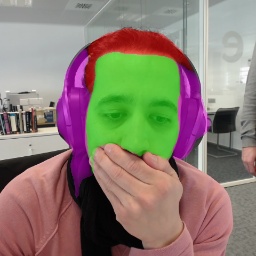}\figcolsep

\end{center}
\captionsep
   \caption{Labeling faces at the lab: occluding with hands. Successful segmentation examples of hands occluding significant parts of the face.}
\label{fig:webcam_occludinghands}
\end{figure}

\begin{figure}[t]
\begin{center}
\includegraphics[width=0.124\linewidth]{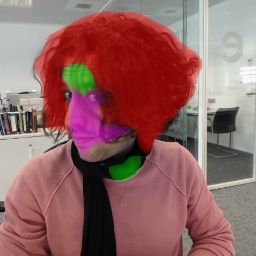}\figcolsep
\includegraphics[width=0.124\linewidth]{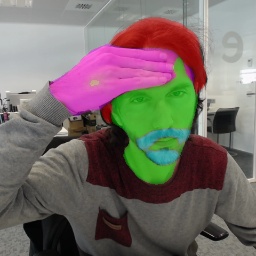}\figcolsep
\includegraphics[width=0.124\linewidth]{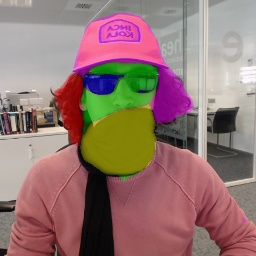}\figcolsep
\includegraphics[width=0.124\linewidth]{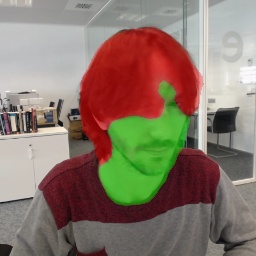}\figcolsep
\includegraphics[width=0.124\linewidth]{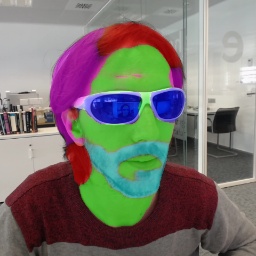}\figcolsep
\includegraphics[width=0.124\linewidth]{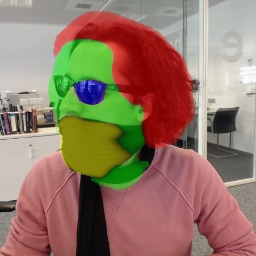}\figcolsep
\includegraphics[width=0.124\linewidth]{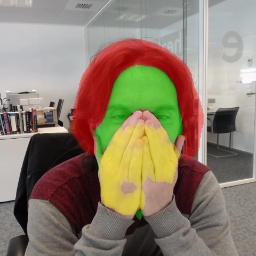}\figcolsep
\includegraphics[width=0.124\linewidth]{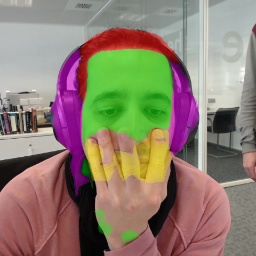}\figcolsep  

\end{center}
\captionsep
   \caption{Labeling faces at the lab: limitations due to severe head rotations, hands at unseen locations, and other category confusion.}
\label{fig:webcam_severalfailures}

\end{figure}

A set of cases with target categories successfully identified is illustrated in Fig.~\ref{fig:webcam_success}. Head-wearables such as caps, wool hats or helmets were properly categorized, sunglasses were identified, and hair was acceptably segmented in a wide range of situations. The last three frames show how the same sunglasses transited from \textit{sunglasses} to \textit{head-wearable} when moving from the eyes to the top of the head.
The mouth-occluding objects were typically designated as \textit{mouth-mask} as shown in Fig.~\ref{fig:webcam_mouthmasks}, which reveals the networks capability to generalize with purely synthetic data.
In Fig.~\ref{fig:webcam_occludinghands} a set of examples are depicted to show the ability of the segmentation network to cope with occluding hands. Note that those cases resembling the artificially augmented assets were properly segmented, even if the major part of the face gets occluded.

Other explanatory failures and limitations are also depicted in Fig.~\ref{fig:webcam_severalfailures}. Some simply happened spuriously, while others do have a direct link to the ELFW's particularities.
For instance, severe rotations lose the face track or beard, and can degenerate to a large hallo-effect. 
Hands were miss-classified as \textit{head-wearable} if placed where wearables typically are expected ---for instance, covering ears or hair---, because no \textit{hand} data was augmented on such upper head locations. 
On its turn, head wearables may harden the classification of objects such as sunglasses. 
Finally, in contrast to Fig.~\ref{fig:webcam_occludinghands}, hands occluding the mouth were sometimes confused with mouth-masks, specially when the nose was occluded too, which again does not frequently occur in the actual data augmentation strategy.

\section{Conclusions}

In this work, the ELFW dataset has been presented, an extension of the widely used LFW dataset for semantic segmentation. It expands the set of images for which semantic ground-truth was available by labeling new images, defining new categories and correcting existing label maps. The main goal was to provide a broader contextual set of classes that are usually present around faces and may particularly harden identification and facial understanding in general. Different category augmentation strategies were deployed, which yielded better segmentation results on benchmarking deep models for the targeted classes, preserving and sometimes improving the performance for the remaining ones. In particular, we have also observed that the segmentation models were able to generalize to classes that were only seen synthetically at the training stage.




%

\newpage

\bibliographystyle{splncs04}
\bibliography{eccv2020elfw}

\end{document}